%% file: main.tex
\documentclass[lettersize,journal]{IEEEtran}
\usepackage[colorlinks,linkcolor=black,anchorcolor=black,citecolor=blue]{hyperref}
\usepackage{amsmath,amsfonts}
\usepackage{algorithmic}
\usepackage{algorithm}
\usepackage{array}
\usepackage[caption=false,font=normalsize,labelfont=sf,textfont=sf]{subfig}
\usepackage{textcomp}
\usepackage{stfloats}
\usepackage{url}
\usepackage{verbatim}
\usepackage{graphicx}
\usepackage{cite}
\usepackage{amssymb}
\usepackage{booktabs}
\usepackage{multirow}
\usepackage{multicol}
\usepackage{makecell}
\begin{document}

\title{
Bidirectional Correlation-Driven Inter-Frame Interaction Transformer for Referring Video Object Segmentation
}

\author{Meng Lan, Fu Rong, Zuchao Li, Wei Yu, Lefei Zhang,~\IEEEmembership{Senior Member,~IEEE}

\thanks{The authors are with the National Engineering Research Center for Multimedia Software, School of Computer Science, Wuhan University, Wuhan, 430072, P. R. China, and also with the Hubei Luojia Laboratory, Wuhan 430079, P. R. China. (e-mail: {menglan, furong, zcli-charlie, yuwei, zhanglefei}@whu.edu.cn). This work was supported by the National Natural Science Foundation of China under Grants 62122060, 62076188, and the Special Fund of Hubei Luojia Laboratory under Grant 220100014. (Corresponding author: Lefei Zhang.)}}%



\maketitle

\begin{abstract}
Referring video object segmentation (RVOS) aims to segment the target object in a video sequence described by a language expression. Typical multimodal Transformer based RVOS approaches process video sequence in a frame-independent manner to reduce the high computational cost, which however restricts the performance due to the lack of inter-frame interaction for temporal coherence modeling and spatio-temporal representation learning of the referred object. Besides, the absence of sufficient cross-modal interactions results in weak correlation between the visual and linguistic features, which increases the difficulty of decoding the target information and limits the performance of the model. In this paper, we propose a bidirectional correlation-driven inter-frame interaction Transformer, dubbed BIFIT, to address these issues in RVOS. Specifically, we design a lightweight and plug-and-play inter-frame interaction module in the Transformer decoder to efficiently learn the spatio-temporal features of the referred object, so as to decode the object information in the video sequence more precisely and generate more accurate segmentation results. Moreover, a bidirectional vision-language interaction module is implemented before the multimodal Transformer to enhance the correlation between the visual and linguistic features, thus facilitating the language queries to decode more precise object information from visual features and ultimately improving the segmentation performance. Extensive experimental results on four benchmarks validate the superiority of our BIFIT over state-of-the-art methods and the effectiveness of our proposed modules. 


\end{abstract}


\section{Introduction}
\IEEEPARstart{R}{eferring} video object segmentation aims to segment the target object in a video sequence described by a language expression \cite{Urvos}. This emerging multimodal task has attracted great attention in the research community since it provides a more natural way for human-computer interaction. RVOS has a wide range of applications, $e.g.$, language-based video editing and surveillance. Compared with the semi-supervised video object segmentation (SVOS) task \cite{EGMN}, which relies on the mask annotations in the first frame \cite{SITVOS}, RVOS is more challenging due to the diversity of language expressions and the difficulties in exploiting the cross-modal knowledge.

Various RVOS datasets and approaches have been proposed in the advancement of the field. For example, Khoreva \textit{et al.} \cite{khoreva2018video} establishes a new benchmark for RVOS by augmenting the popular VOS benchmark, $i.e.$, DAVIS2017 \cite{DAVIS2017}, with language descriptions. Besides, a baseline method is provided by combining the referring expression grounding model \cite{yu2018mattnet} and the segmentation model. URVOS \cite{Urvos} constructs the first large-scale RVOS dataset called Refer-Youtube-VOS by annotating the referring expressions for the Youtube-VOS dataset \cite{ytvos}. Recently, the multimodal Transformer based RVOS methods \cite{MTTR,Referformer} are drawing increasing attention for their impressive performance. They formulate RVOS as a sequence prediction task and extent the DETR architecture \cite{DETR} to generate predictions for all possible objects in the video sequence prior to selecting the one that matches the language description. Among them, ReferFormer \cite{Referformer} leverages the language description as the query of the multimodal Transformer decoder and produces more accurate instance embeddings of the referred object for the final instance sequence prediction, which achieves state-of-the-art performance.

\begin{figure}[t]
    \centering
     \includegraphics[width=\linewidth]{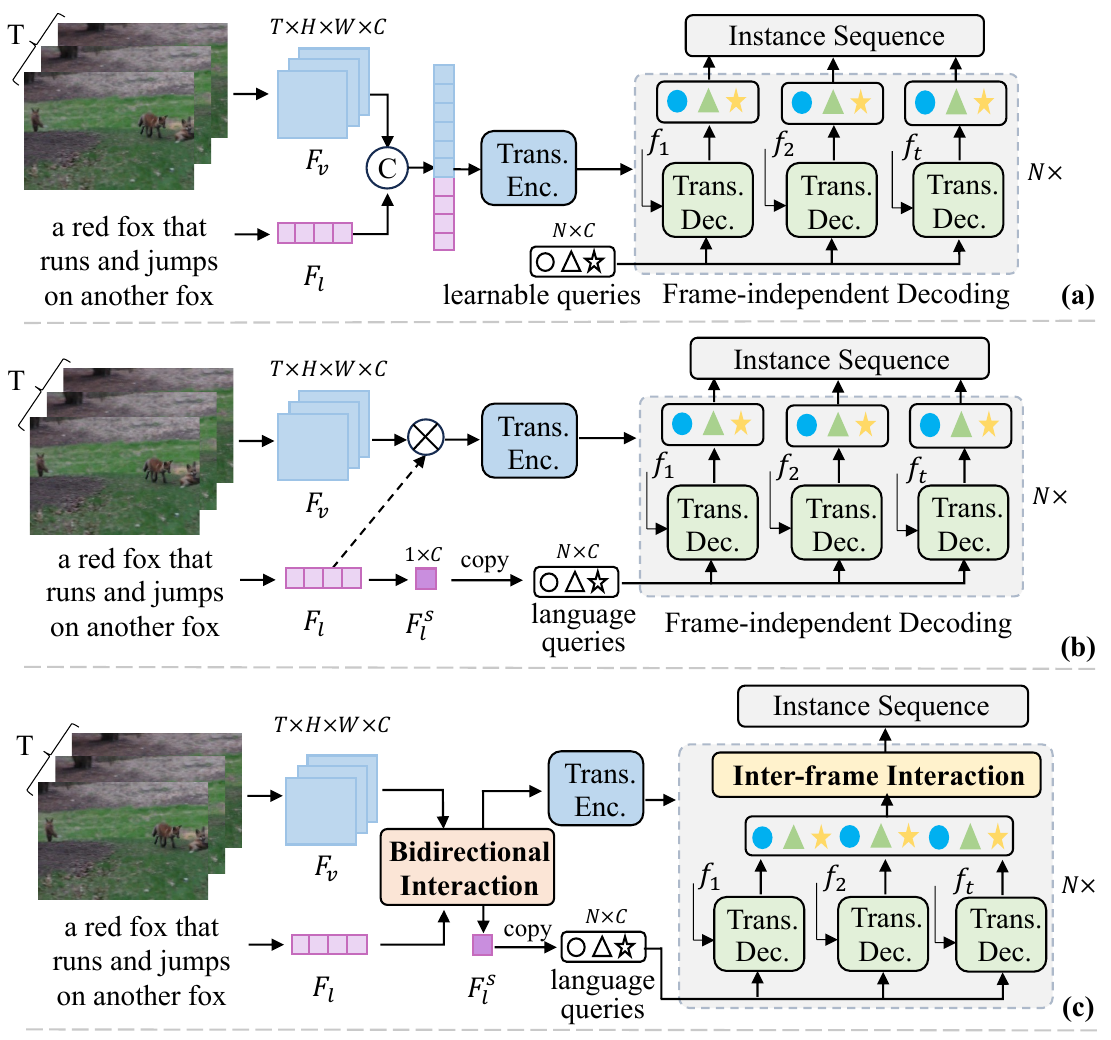}
     \caption{Schematic of the framework comparison between the proposed model and previous multimodal Transformer based RVOS methods. (a) MTTR \cite{MTTR}. (b) ReferFormer \cite{Referformer}. (c) Ours (BIFIT)}
     \vspace{-3mm}
     \label{fig1}
\end{figure}

Despite the impressive performance of the multimodal Transformer based RVOS methods on various benchmarks, we contend that there is room for further improvement in two key areas. \textit{\textbf{First}}, traditional methods are rudimentary and inadequate in terms of cross-modal fusion. As shown in Fig.\ref{fig1} (a), to adapt the RVOS task to the DETR architecture \cite{DETR}, the pioneering MTTR \cite{MTTR} simply concatenates the linguistic and visual features and then feeds them into the Transformer. However, the simple concatenation ignores the properties inherent in each modal feature and may limit effective interactions between cross-modal features. Therefore, the widely used cross-modal attention mechanism \cite{lin2021structured,yang2022object,ding2023bilateral} is applied in ReferFormer \cite{Referformer}, where attention based cross-modal interaction is implemented prior to the multimodal Transformer, as shown in Fig.\ref{fig1} (b). Nevertheless, ReferFormer only performs the unidirectional text-guided visual feature enhancement, and directly adopts the raw and high-level sentence feature without any interaction with visual features as the language queries to decode the image features. We argue that the weak correlation between visual and linguistic features before multimodal Transformer exacerbates the challenge of accurately decoding the object information using language queries, and limits the further improvement of model performance. Consequently, how to efficiently build strong correlation between the linguistic and visual features is an essential strategy for advancing this field. \textit{\textbf{Second}}, to mitigate the computational burden, both MTTR and ReferFormer retrieve the video sequence in a frame-independent manner and the cross-frame correlation relies heavily on the sharing input queries, as depicted in Fig.\ref{fig1}. However, this operation may result in performance degradation due to the lack of explicit inter-frame interaction in the decoder, which is crucial for capturing temporal coherence and learning spatio-temporal representations of the referred object. Hence, efficiently learning the spatio-temporal representation of the target object is another indispensable strategy for further improvement.

In this study, we propose BIFIT, a bidirectional correlation-driven inter-frame interaction Transformer for RVOS, to accomplish the aforementioned strategies, as illustrated in Fig.\ref{fig1} (c). First, based on the multimodal Transformer architecture, we design an \textit{inter-frame interaction module} for the Transformer decoder. Different from the complex temporal coherence modeling ways in video instance segmentation task, our module is simple and can efficiently capture the temporal coherence and learn the spatio-temporal representation of the referred object in a plug-and-play manner. Specifically, an inter-frame interaction layer is inserted after each decoder layer in the Transformer. This layer initially unfolds the low-dimensional instance embeddings generated by the frame-independent decoding in the spatio-temporal dimension, and then learns inter-frame global correlation and spatio-temporal representation of queried object through self-attention mechanism. The instance embeddings output from the inter-frame interaction layer are converted back to the frame-independent state and then fed into the next decoder layer or directly outputted from the decoder. In this manner, the multimodal Transformer could decode more coherent object information across the video frames and lead to more accurate segmentation results.

Subsequently, we further develop a \textit{bidirectional vision-language interaction module} and integrated it prior to the multimodal Transformer in order to reinforce the cross-modal correlation between the visual and linguistic features, thereby facilitating the language queries to decode more precise object information from visual features in the Transformer decoder. Concretely, the bidirectional vision-language interaction module maintains two parallel submodules that leverage the cross-modal features to enhance visual features and linguistic features, respectively. The strong correlation between the cross-modal features could facilitate of the language queries based decoding process and promote the model to decode more accurate instance embedding, thereby improving the segmentation performance of the model.

Empirical results on four benchmark datasets demonstrate that these two straightforward and effective modules can significantly improve the performance of multmodal Transformer based RVOS methods and enable the BIFIT model to achieve state-of-the-art resutls. The main contributions of this work can be summarized as follows:
\begin{itemize}
\item We propose a bidirectional correlation-driven inter-frame interaction Transformer (BIFIT) framework for RVOS, which aims to efficiently decode precise and consistent instance embeddings from video sequence by enhancing the correlation between the cross-modal features and learning the spatio-temporal target representation. BIFIT outperforms the previous cutting-edge methods on several benchmarks and realizes state-of-the-art performance.

\item We design an inter-frame interaction module for the Transformer decoder to efficiently model the temporal coherence and learn the spatio-temporal representation of the queried object, so as to decode more consistent instance embeddings for predicting high-quality segmentation results.

\item We develop a bidirectional vision-language interaction module before the multimodal Transformer to boost the correlation between the cross-modal features, thus facilitating the language queries to decode more precise object information from visual features and further improving the model performance.

\end{itemize}

\section{Related work}

\subsection{Referring Video Object Segmentation}
RVOS task poses a greater challenge compared to the SVOS, as it solely relies on the language description rather than the object mask as object reference. RVOS can be regarded as an extension of referring image segmentation (RIS) \cite{cris} by extending the input from the image domain to the video domain. Therefore, an intuitive approach for RVOS is applying the RIS methods on the video frames independently, $e.g.$, RefVOS \cite{bellver2020refvos}. However, such an approach fails to consider the temporal information across frames, resulting in inconsistent target predictions due to the scene and object appearance variations. To solve this problem, URVOS \cite{Urvos} treats this task as a joint problem of RIS in an image and mask propagation in a video. It proposes a unified framework that includes a memory attention module to propagate the target information to the current frame. To learn a more effective target representation, VTCapsule \cite{mcintosh2020visual} encodes each modality in capsules while ACGA \cite{wang2019asymmetric} designs an asymmetric cross-guided attention network to enhance the linguistic and visual features. YOFO \cite{yofo} extends the online learning based architecture in SVOS \cite{LLB} to RVOS by introducing a multi-scale cross-modal feature mining block. To capture the spatial-temporal consistency of the referred object, LBSTI \cite{ding2022language} proposes to utilize language as an intermediary bridge to accomplish explicit and adaptive spatial-temporal interaction. MLSA \cite{wu2022multi} integrates multi-level target features to enable more effective vision-language semantic alignment. Inspired by the success of the query-based Transformer frameworks in other fields \cite{DETR,VisTR}, query-based multimodal Transformers are also explored in the RVOS task. MTTR \cite{MTTR} models the RVOS task as a sequence prediction problem and processes the video and text together in a multimodal Transformer. ReferFormer \cite{Referformer} follows this idea while using the sentence feature as the language queries to find the referred object within the Deformable-DETR \cite{zhu2020deformable}. OnlineRefer \cite{OnlineRefer} proposes a simple yet effective online model using explicit query propagation to achieve temporal association. Additionally, R$^{2}$-VOS \cite{r2vos} proposes the relational cycle consistency constraint to enhance the semantic alignment between visual and textual modalities, thus improving the segmentation accuracy.

\begin{figure*}[t]
    \centering
     \includegraphics[width=0.94\linewidth]{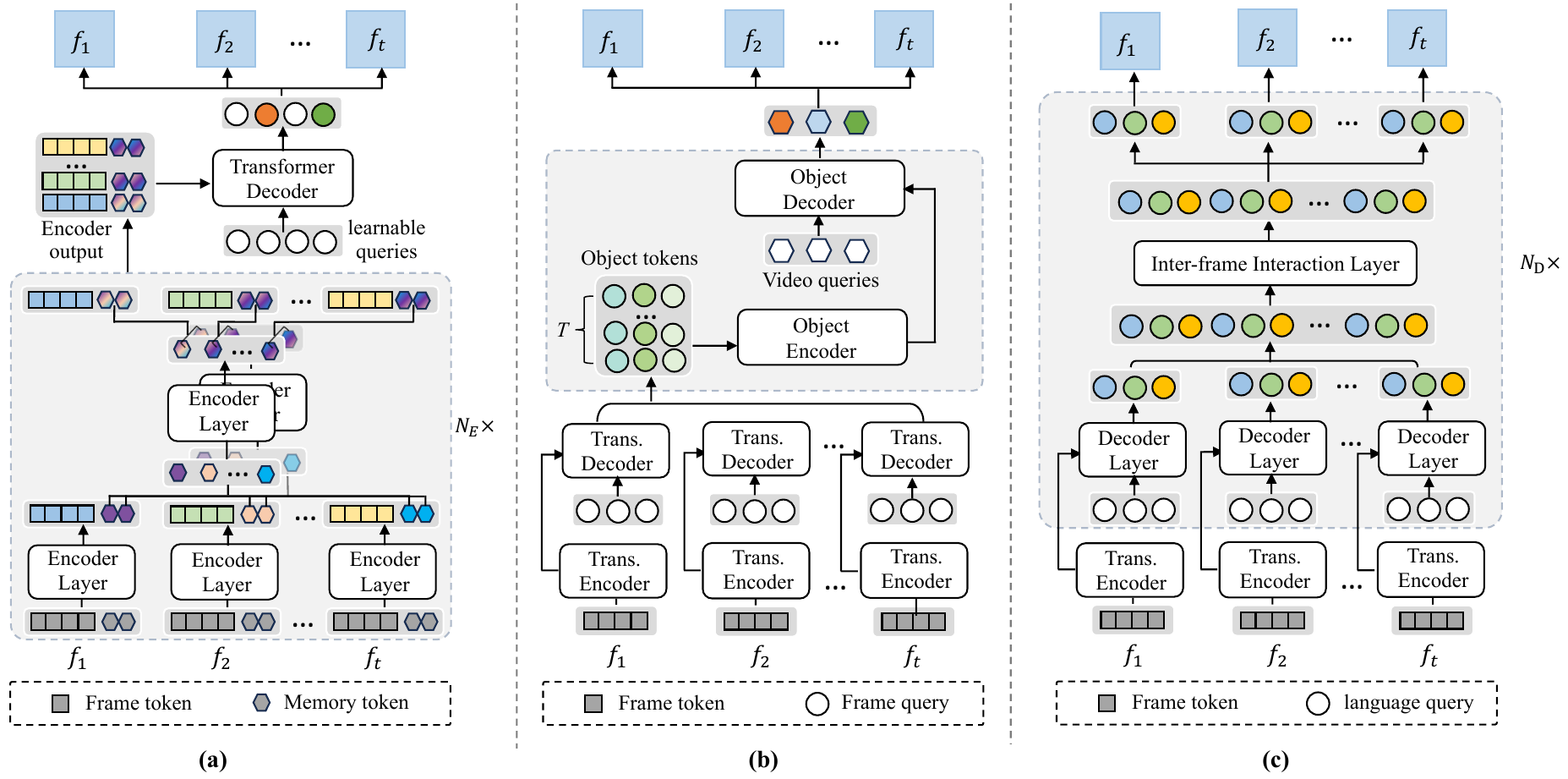}
     \caption{Comparison of various approaches to learning target temporal coherence in video sequences. (a) IFC: Learning in encoder \cite{ifc}. (b) VITA: Learning in individual Transformer \cite{vita}. (c) Ours BIFIT: Learning in decoder.}
     \vspace{-2mm}
     \label{ifi_ways}
\end{figure*}

\vspace{-2mm}
\subsection{Video Instance Segmentation}
Video instance segmentation (VIS) \cite{qin2021learning} could provide inspiration for RVOS task, since VIS intends to segment all seen objects and RVOS can be regarded as a special case of it, $i.e.$, segmenting referred object. DETR \cite{DETR} is a widely used architecture in object detection field, which uses a set of object queries to infer the global context of the image and the relationships between the objects, and then outputs a set of predicted sequences in parallel. The idea is also introduced to the VIS task. VisTR \cite{VisTR} introduces the DETR model to VIS by treating the VIS as an end-to-end parallel sequence prediction problem and using parallel sequence decoding to solve it. However, VisTR integrates the spatio-temporal dimension of the video features and feeds them directly into the Transformer, resulting in a huge computational burden and limited instance segmentation performance. To solve this problem, as shown in Fig.\ref{ifi_ways} (a), IFC \cite{ifc} proposes the inter-frame communication Transformers, which incorporates memory tokens in the Transformer encoder to store the overall context of the video clip and then queries the linked frame and memory tokens independently in the decoder. The memory tokens reduce the overhead of spatio-temporal information transfer in the encoder and introduces the global information for subsequent decoding of each frame. Subsequently, VITA \cite{vita} proposes a two-stage strategy to progressively learn the spatio-temporal representation of the objects. As presented in Fig.\ref{ifi_ways} (b), VITA uses the mask2former model \cite{mask2former} to distill object-specific contexts into object tokens, and then accomplishes the video-level understanding by associating frame-level object tokens in individual Transformer, where the generated video-level instance embeddings guide the instance segmentation of all the frames. Inspired by these approaches in learning the spatio-temporal representation of the objects, we propose a new and straightforward inter-frame interaction mechanism for RVOS task that directly performs the inter-frame interaction on candidate object queries of different frames during the decoding process of the Transformer decoder. This strategy not only efficiently realizes spatio-temporal representation learning of the referred object across video frames but also significantly improve the segmentation performance.


\begin{figure*}[t]
    \centering
     \includegraphics[width=0.94\linewidth]{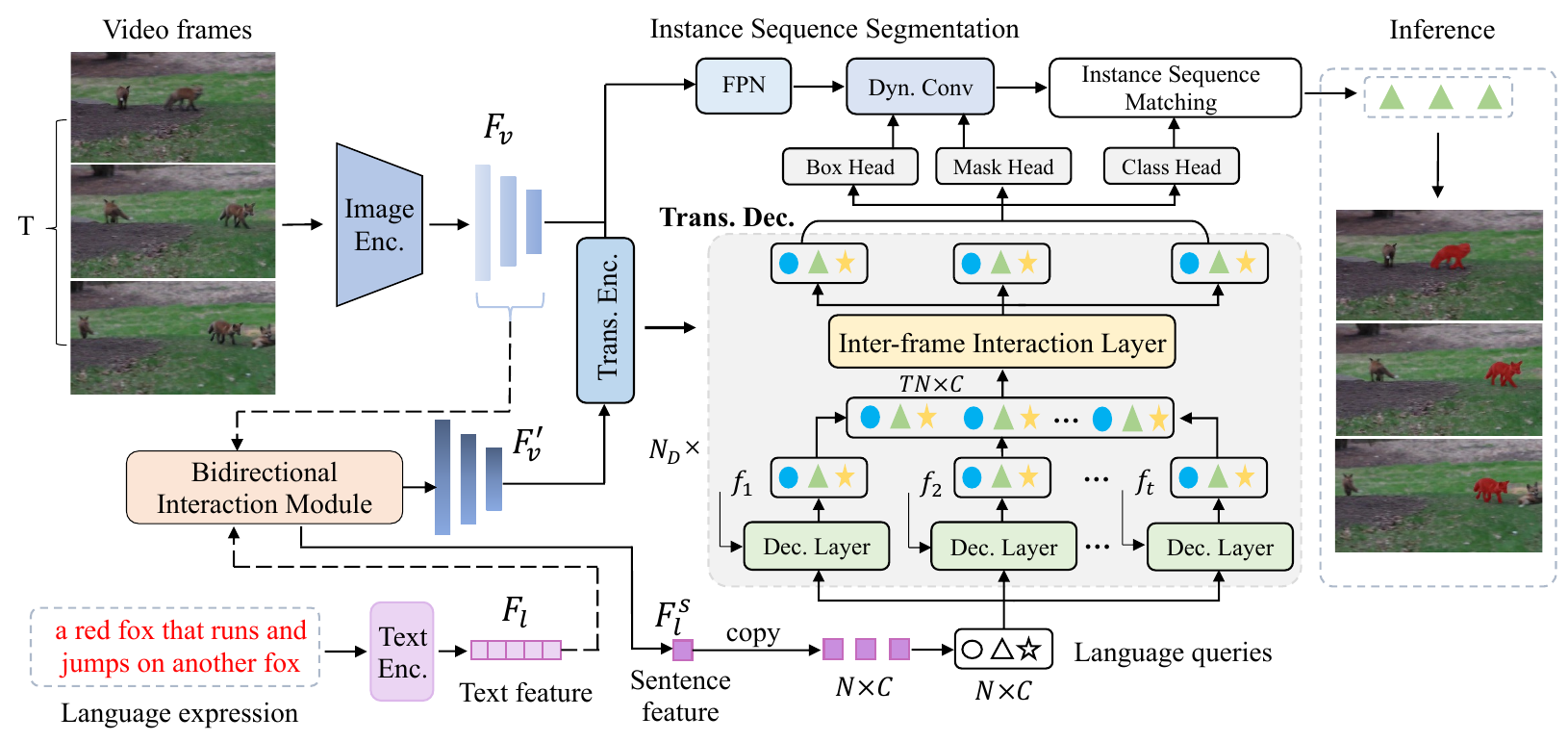}
     \caption{An overview of the proposed BIFIT method. It mainly consists of four parts: the image and text encoders, the bidirectional vision-language interaction module, the multimodal Transformer with inter-frame interaction module and the instance sequence segmentation part. The bidirectional vision-language interaction module establishes a strong correlation between the visual and linguistic features before they are sent into the multimodal Transformer. The inter-frame interaction module in the Transformer decoder enables the instance embeddings for each frame with temporal coherence and spatio-temporal representation. $\{f_t\}_{t=1}^{T}$ is the output of Transformer encoder. Here the same colors and shapes in the queries refer to the same object in different frames.}
     \label{framework}
\end{figure*}

\section{Method}
\subsection{Overview}
The overview of our proposed BIFIT is illustrated in Fig.\ref{framework}. It mainly consists of four parts: the image and text encoders, the bidirectional vision-language interaction module, the multimodal Transformer with inter-frame interaction module and the instance sequence segmentation part. During inference, given a video sequence $\mathcal{V}=\left\{I_t\right\}_{t=1}^T$ with $T$ frames and a referring expression of the target object $\mathcal{E}=\left\{e_l\right\}_{l=1}^L$ with $L$ words, the image and text encoders first extract the multi-level visual features of the $T$ frames and the word feature of the language expression, which are then fed into the bidirectional vision-language interaction module to produce language-enhanced visual features and the vision-enhanced sentence feature. The Transformer encoder takes the enhanced multi-level visual features as input and its outputs along with the sentence feature are submitted to the Transformer decoder, where the sentence feature serves as the language queries to decode the object information from the frame features and generates the instance embeddings. The inter-frame interaction module inserted in the Transformer decoder enables the instance embeddings for each frame with temporal coherence and spatio-temporal representation. Finally, the instance sequence segmentation part integrates the instance embeddings and the visual features to predict accurate segmentation sequence for referred object.

\subsection{Feature Extraction}
\subsubsection{Image Encoder} For the frames in the video sequence, we adopt the image encoder to extract the multi-level visual features of each frame independently and obtain the visual feature sequence $F_{v}=\{F_t\}_{t=1}^{T}$, where $F_t$ denotes the multi-level features for the $t$-th frame. Specifically, $F_t$ is a four-level pyramid features, in which the first three-level features are the last three stage features of the image encoder with spatial strides of $\{8,16,32\}$, and the last-level feature is obtained by downsampling the 32-stride feature using a convolutional layer with stride 2, thus $F_t$ is the four-level pyramid features with strides of $\{8,16,32,64\}$. 


\subsubsection{Text Encoder} The linguistic feature are extracted from the language expression using the off-the-shelf text encoder, called RoBERTa \cite{liu2019roberta}. Different from ReferFormer \cite{Referformer}, as illustrated in Fig.\ref{fig1} (b), that generates the word-level text feature (text feature) for cross-modal fusion and the sentence-level feature (sentence feature) for language queries, we only need the word feature $F_{l}\in R^{{L}\times{C}}$, which contains the feature embedding of each word in the language expression. Due to its semantic richness and inclusion of more object information, the text feature is better suited for performing fine-grained cross-modal interactions with visual features to generate vision-enhanced text feature and sentence feature.

\subsection{Bidirectional Vision-Language Interaction Module}
For the task of RVOS, which involves segmenting a specified object in a given video sequence based on language expression, it is crucial to align cross-modal features and transfer semantics from the language modality to the visual modality for accurate target object localization and segmentation.

The pioneering MTTR \cite{MTTR} chooses to form a simple cross-modal fusion feature by concatenating visual and linguistic features, which is then fed into Transformer for sequence prediction. In contrast, ReferFormer proposes to use the sentence feature of the language expression as the queries to iteratively interact with visual features in the Transformer decoder and generate instance embeddings that contain representation information of the referred object. Although ReferFormer has performed the unidirectional text-guided visual feature enhancement based on cross-attention, the raw sentence feature that serves as the language queries is abstract and contains only high-level semantic information. Furthermore, its correlation with the visual features is weak due to their lack of close interaction before the Transformer. We argue that the weak correlation between high-level sentence feature and the visual features will increases the difficulty of decoding the object information accurately in the Transformer decoder, ultimately constraining the model performance.

\begin{figure}[t]
    \centering
     \includegraphics[width=0.98\linewidth]{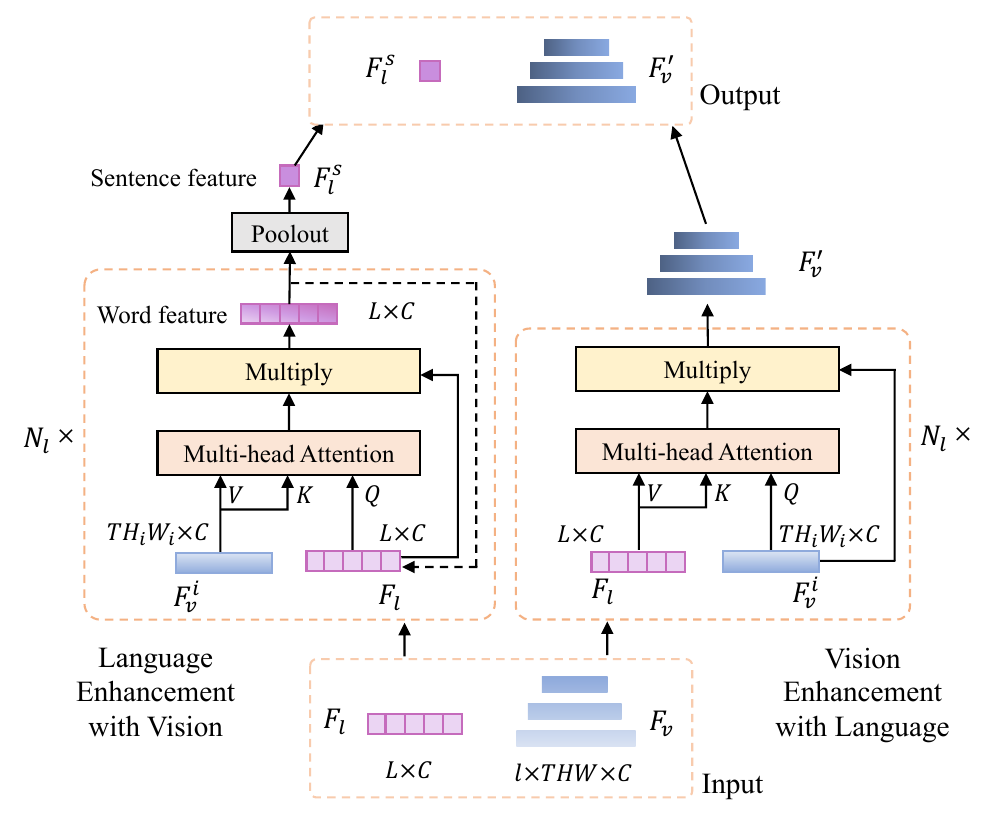}
     \caption{The structure of bidirectional vision-language interaction module.}
     \vspace{-2mm}
     \label{VL-enhance}
\end{figure}

Therefore, in this paper, we devise the \textit{bidirectional vision-language interaction module} to reinforce the cross-modal correlation between the visual and text features, and place it before the multimodal Transformer. As shown in Fig.\ref{VL-enhance}, the bidirectional interaction module is composed of two parallel submodules, namely Vision Enhancement with Language (VEwL) submodule and Language Enhancement with Vision (LEwV) submodule. Both submodules take the raw text feature $F_{l}$ and multi-level visual feature sequence $F_{v}$ as input, and the VEwL submodule outputs the language-enhanced multi-level visual features $F_{v}^{'}$ while the LEwV submodule produces the vision-enhanced sentence feature $F_{l}^{s}$. We will describe these two submodules in detail in the following part.

\subsubsection{Language Enhancement with Vision Submodule} LEwV submodule starts with a cross-modal interaction between text feature and visual features, where the visual features serve as the guidance to enhance the text feature. As shown in the left part of Fig.\ref{VL-enhance}, the raw text feature $F_{l}$ and the original multi-level visual feature sequence $F_{v}$ are the input of the LEwV submodule, which is composed of a multi-head attention layer and a multiplier layer \cite{yang2022lavt}. LEwV submodule iteratively processes the text feature and the single-level visual feature sequence for $N_{l}$ times until the text feature interacts with all levels of the multi-level visual feature sequence. $N_{l}$ denotes the number of levels of the multi-level visual feature sequence $F_{v}$.

The multi-head attention layer adopts the cross-attention mechanism to accomplish the cross-modal interaction between the text and visual features, where the text feature $F_{l} \in R^{{L}\times{C}}$ serves as the query (Q) and the single-level visual features $F_{v}^{i} \in R^{{TH_{i}W_{i}}\times{C}} $ is used as the key-value pair (K-V). Here $H_{i}$ and $W_{i}$ are the height and width of the $i$-th level of features, and $C$ is the channel number. For simplicity, we present the cross-attention based cross-modal interaction process in a single-head manner. For the $i$-th iteration, the attention layer first obtains the cross-modal similarity matrix $A_{lv}^{i} \in R^{L \times {TH_{i}W_{i}}}$ by computing the similarity between each word embedding and each pixel embedding as follows:
\begin{equation}
A_{lv}^{i}=Softmax(\frac{F_{l}^{i}W^{Q}\cdot{(F_{v}^{i}W^{K})^{T}}}{\sqrt{d_{k}}}),
\end{equation}
where $W^{Q}, W^{K}\in R^{C \times {d_{k}}}$ are learnable linear projections. Then we use $A_{lv}^{i}$ to aggregate the language-related target information in the visual features and multiply with the input text feature $F_{l}^{i}$ to obtain the vision-enhanced text feature $F_{l}^{i+1}$:
\begin{equation}
F_{l}^{i+1}=(A_{lv}^{i}F_{v}^{i}W^{V})\cdot F_{l}^{i},
\end{equation} 
where $W^{v}$ is the learnable linear projection. To obtain positional information, a fixed two-dimensional sinusoidal positional encoding is added to the visual features before the cross-attention process. The raw text feature $F_{l}$ is $F_{l}^{0}$ in the first iteration.

After the last iteration of the cross-modal interaction, the final vision-enhanced text feature $F_{l}^{'}$ is converted to vision-enhanced sentence feature $F_{l}^{s} \in \mathbb{R}^{1 \times C}$ by performing the \textit{poolout} operation in the RoBERTa model.

\subsubsection{Vision Enhancement with Language Submodule} VEwL submodule has the same submodule architecture and input features with the LEwV submodule, as presented in the right part of Fig.\ref{VL-enhance}. However, the VEwL submodule has different running procedure. Each level of the original multi-level visual feature sequence $F_{v}^{l}$ interacts with the raw text feature $F_{l}$ individually via the multi-head attention layer and the multiplier layer. Similarly, for each cross-modal interaction, the multi-head attention layer performs the cross-attention on the text feature $F_{l}$ and the single-level visual feature sequence $F_{v}^{i}$, where the visual features act as the query (Q) and the text feature serves as the key-value pair (K-V). The attention layer first derives the cross-modal similarity matrix $A_{vl}^{i}\in R^{{TH_{i}W_{i}}\times{L}}$ by computing the similarity between each pixel embedding and each word embedding as follows:
\begin{equation}
A_{vl}^{i}=Softmax(\frac{F_{v}^{i}W^{Q}\cdot{(F_{l}W^{K})^{T}}}{\sqrt{d_{k}}}),
\end{equation}
where $W^{Q}, W^{K}\in R^{C\times{d_{k}}}$ are learnable linear projections. Then the similarity matrix is used to aggregate the vision-related object information in the text features followed by the multiplication with the input visual feature $F_{v}^{i}$ to obtain the language-enhanced single-level visual feature $F_{v}^{i'}$
\begin{equation}
F_{v}^{i'}=A_{vl}^{i}F_{l}W^{V}\cdot F_{v}^{i},
\end{equation}
where $W^{V}$ is the learnable linear projection. To obtain positional information, we add fixed one-dimensional sinusoidal positional encoding to the text feature before the cross-attention operation.

After the cross-modal interaction of each level of the multi-level visual feature sequence $F_{v}$ with the raw text features $F_{l}$, we obtain the language-enhanced multi-level visual feature sequence $F_{v}^{'}$.

\subsection{Multimodel Transformer}
The multimodal Transformer aims to exploit the vision-enhanced sentence feature and the language-enhanced multi-level visual feature sequence to produce the target-aware instance embeddings, which are converted to conditional convolution kernels to perform conditional convolution \cite{tian2020conditional} on the visual features and generate the final segmentation masks. Here we adopt the Deformable-DETR \cite{zhu2020deformable} as the multimodal Transformer like \cite{Referformer} and use the vision-enhanced sentence feature $F_{l}^{s}$ as the language queries of the decoder to find the referred object.

\subsubsection{Transformer Encoder} Before feeding the language-enhanced multi-level visual features into the Transformer encoder, the fixed 2D positional encodings are added to the feature maps of each frame to reinforce the position information. After that, the encoder processes these multi-level features in a frame-independent manner, and the resulting output features are then fed into the decoder. Besides, the first three stage features of Transformer encoder output and the backbone feature with spatial stride of 4 are sent together into the cross-modal FPN \cite{Referformer} to generate the final feature maps for segmentation, $i.e.$, $F_{seg} = \left \{ f_{\text{seg}}^{t} \right \}_{t=1}^{T}$, where $f_{\text{seg}}^{t} \in \mathbb{R}^{\frac{H}{4} \times \frac{W}{4} \times C}$, $H$ and $W$ are the height and width of the input frames.

\subsubsection{Transformer Decoder.} As shown in Fig.\ref{framework}, to enhance the feature learning ability of the decoder, we repeat the sentence feature for $N$ times to generate $N$ object queries for each frame, following the strategy in \cite{Referformer}. The encoder output, the language queries, and the learnable reference point embeddings as in Deformable-DETR are fed into the decoder. Then, the language queries interact with the visual features and try to find the referred object only, resulting in the set of $N_{q}=T\times N$ instance embeddings. 

Similar with \cite{Referformer}, the decoding process of the language queries and visual features is implemented in a frame-independent fashion, which however leads to a lack of inter-frame communication among the instance embeddings generated for each frame and the absence of temporal coherence of the target object, thereby impacting the final segmentation performance. To tackle this issue and introduce the spatio-temporal representation for the instance embeddings, we propose an inter-frame interaction module that enables the instance embeddings of each frame to take good advantage of the temporal information between frames, allowing for better tracking and segmentation of the referred object in the video sequence. 

\begin{figure}[t]
    \centering
     \includegraphics[width=0.65\linewidth]{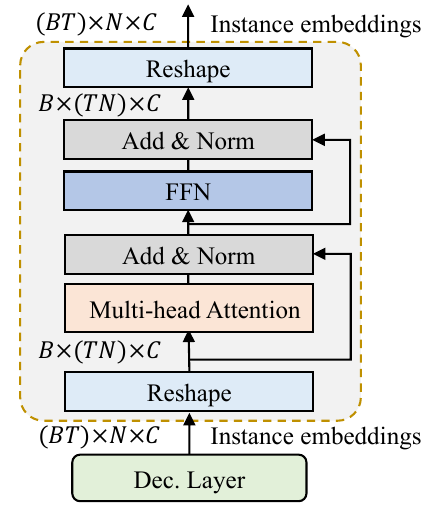}
     \caption{The architecture of the inter-frame interaction layer within the inter-frame interaction module.}
     \vspace{-2mm}      
     \label{interaction}
\end{figure}

\subsubsection{\textbf{Inter-frame Interaction Module}} Inter-frame interaction module is a lightweight and plug-and-play module for the multimodal Transformer decoder. The module contains several inter-frame interaction layer, each of which is inserted behind each decoder layer of the Transformer to efficiently model the temporal coherence and learn the spatio-temporal representation for the instance embeddings, as shown in Fig.\ref{interaction}. Specifically, the instance embeddings $Q \in R^{(BT)\times N \times C}$ generated by the frame-independent decoding process are first unfolded in the spatio-temporal dimension to obtain $Q \in R^{B \times (TN) \times C}$. Here $B$ is the batch size. Then, instance embeddings $Q$ are fed into the standard multi-head self-attention layer and the feed-forward network (FFN) \cite{attention}, where the inter-frame global correlation and spatio-temporal representation of the queried object are learned. The instance embeddings output from the inter-frame interaction layer are transformed back into the frame-independent state, $i.e.$, $Q \in R^{(BT)\times N \times C}$, and then sent to the next decoder layer or directly as the output of the decoder. The inter-frame interaction layer could be formulated as follows:
\begin{equation}
\begin{aligned}
    Q_1 &= LN(Atten(Re(Q))+Re(Q) ), \\
    Q_2 &= Re(LN(FFN(Q_1)+Q_1)),
\end{aligned}
\end{equation}
where $Re(\cdot)$ represents the reshape operation, $Atten(\cdot)$ represents the multi-head attention layer, $LN(\cdot)$ denotes the layer normalization, and $FFN(\cdot)$ denotes the feed-forward network.

Here, we analyze the complexity of the added inter-frame interaction module. The complexity of each inter-frame interaction layer is $\mathcal{O}\left(C^2 (TN)+C (TN)^2\right)$. Since $N$ is kept small ($e.g.$, 5) and $T$ is constrained to a maximum of 40 due to the hardware limitation, the computation needed for inter-frame interaction layer almost could be neglected.

\subsection{Instance Sequence Segmentation}
As shown in \ref{framework}, three prediction heads are built on top of the decoder to further transform the $N_{q}$ instance embeddings output from the decoder, $i.e.$, box head, mask head, and class head. The class head predicts whether the predicted instance is described by the expression or whether the instance is available in the current frame. The mask head consists of three linear layers and is responsible for predicting the parameters of the conditional convolution kernels $\Omega={\{\omega_{i}\}}_{i=1}^{N_{q}}$, which are reshaped to form three $1\times 1$ convolution kernels. The box head is a 3-layer feed-forward network with ReLU activation except for the last layer. It predicts the box location of the referred object. Finally, we implement the instance sequence segmentation and produce the final frame-order mask sequence predictions $S\in{R^{T\times N\times \frac{W}{4}\times \frac{H}{4}}}$ by applying the conditional convolution kernels $\Omega={\{\omega_{i}\}}_{i=1}^{N_{q}}$ on the corresponding feature maps, which are the concatenation of the feature maps $F_{seg}$ and relative box coordinates as \cite{Referformer} did. 

During the training process, the predicted instance sequence is treated as a whole and supervised by the instance matching strategy \cite{VisTR}. We denote the instance prediction sequences as $\hat{y}=\left\{\hat{y}_i\right\}_{i=1}^N$, and the predictions for the $i$-th instance is denoted as:
\begin{equation}
\hat{y}_i=\left\{\hat{p}_i^t, \hat{b}_i^t, \hat{s}_i^t\right\}_{t=1}^T,
\end{equation}
where $\hat{p_{i}}^{t}\in R^{1}$ is the probability score predicted by the class head for the $t$-th frame in the video sequence. $\hat{b}_{i}^{t}\in R^{4}$ is the normalized coordinates that defines the center point as well as the height and width of the prediction box. $\hat{s}_{i}^{t}\in R^{{\frac{H}{4}}\times\frac{W}{4}}$ is the predicted binary segmentation mask.

The ground truth instance sequence could be represented as $y={\{c^t, b^t, s^t\}}_{t=1}^{T}$, where $c^t$ is an one-hot value that equals 1 when the ground truth instance is visible in the $t$-th frame and 0 otherwise. $b^t$ and $s^t$ are the corresponding normalized box coordinates and segmentation mask. To train the network, we first locate the best prediction sequence from all the instance prediction sequences as the positive sample by minimizing the following matching cost:
\begin{equation}
    \hat{y}_{\text {pos }}=\underset{\hat{y}_i \in \hat{y}}{\arg \min } \mathcal{L}_{\text {match }}\left(y, \hat{y}_i\right),
\end{equation}
where 
\begin{equation}
    \begin{aligned}
\mathcal{L}_{\text {match }}\left(y, \hat{y}_i\right) &=\lambda_{cls} \mathcal{L}_{cls}\left(c, \hat{p}_i\right)+\lambda_{box} \mathcal{L}_{box}\left(b, \hat{b}_i\right) \\
&+\lambda_{mask} \mathcal{L}_{mask}\left(s, \hat{s}_i\right),
    \end{aligned}
\end{equation}
here, $\mathcal{L}_{cls}$ is the focal loss \cite{lin2017focal}, $\mathcal{L}_{cls}$ is the sum of the L1 loss and GIoU loss, and $\mathcal{L}_{mask}$ is the combination of DICE loss \cite{milletari2016v} and binary mask focal loss. The matching cost is calculated from each frame and normalized by the frames number. Then the whole model is optimized by minimizing the matching loss of the positive sample.

During inference, given a video sequence and the language expression, BIFIT could predict $N$ instance sequences corresponding to the $N$ queries. For each prediction sequence, we average the predicted class probabilities over all the frames and get the probability score set $P = {\{p_{i}\}}_{i=1}^{N}$, and we select the sequence with the highest score as the final predictions of the input video sequence without any post-process technique.

\section{Experiences}
\subsection{Datasets and Metrics}
\subsubsection{Datasets} The experiments are conducted on the four popular RVOS benchmarks: Ref-Youtube-VOS \cite{Urvos}, Ref-DAVIS17 \cite{khoreva2018video}, A2D-Sentences and JHMDB-Sentences \cite{gavrilyuk2018actor}. Ref-Youtube-VOS is a large-scale benchmark which covers 3471 videos with 12913 expressions in the training set and 202 videos with 2096 expressions in the validation set. Ref-DAVIS17 is split into 60 videos and 30 videos for training and validation, respectively. We only use the validation set for evaluation. A2D-Sentences and JHMDB-Sentences are built by augmenting the original A2D \cite{xu2015can} and JHMDB \cite{jhuang2013towards} datasets with additional textual annotations. A2D-Sentences consists of 3782 videos, each containing 3-5 frames that are annotated with pixel-level segmentation masks. JHMDB-Sentences comprises a total of 928 videos and their corresponding sentences.

\subsubsection{Evaluation Metrics} Following the standard evaluation protocol \cite{Urvos}, we use the region similarity $J$, contour accuracy $F$, and their average value $J\&F$ for the evaluation on the Ref-Youtube-VOS and Ref-DAVIS17 val sets. Since there is no publicly available ground truth annotations of the Ref-Youtube-VOS val set, we submit our predictions to the official server to get the evaluation results. For A2D-Sentences and JHMDB-Sentences datasets, the model is evaluated in the metrics of Precision@K, Ovrall IoU, Mean IoU and mAP over 0.50:0.05:0.95 like \cite{Referformer}.

\begin{table}[t]
\tabcolsep 9pt
\caption{Comparison with state-of-the-art methods on Ref-Youtube-VOS val set.}
    \begin{center}
        \input{table/refytvos}
    \end{center}
    \label{refytvos}
\end{table}

\begin{table}[t]
\tabcolsep 9pt
\caption{Comparison with state-of-the-art methods on Ref-DAVIS17 val set. $*$ means no fine-tuning on the Ref-DAVIS17 training set.}
    \begin{center}
        \input{table/davis}
    \end{center}
    \vspace{-3mm}
    \label{refdavis}
\end{table}

\subsection{Implementation Details}
\subsubsection{Model Settings} Due to the limitation of our GPUs (four \textbf{24G RTX 3090} for ours $vs$ eight \textbf{32G Tesla V100} for ReferFormer \cite{Referformer} ), we can only use the ResNet50 \cite{he2016deep} pre-trained on ImageNet \cite{deng2009imagenet} as the image encoders for both training and inference stages, ensuring a fair comparison with ReferFormer \cite{Referformer}. For the Multimodal Transformer, we adopt 4 encoder layers and 4 decoder layers with the hidden dimension $C=256$. The number of language queries $N=5$ and the number of the levels of the multi-level visual features $N_l=4$.

\subsubsection{Training Details} The training of our model is divided into two stages as \cite{Referformer}. We first pre-train our BIFIT on the RIS datasets, including Ref-COCO \cite{refcoco}, Ref-COCOg \cite{refcoco}, and Ref-COCO+ \cite{refcocop}, with the number of frames $T=1$ and a batch size of 2 on each GPU. The model is pre-train for 10 epochs with the learning rate reduced by a factor of 0.1 at the 6th and 8th epochs. After the pre-training stage, we employ different fine-tuning strategies to tune the model on different RVOS training sets. 

For the Ref-Youtube-VOS and Ref-DAVIS17 datasets, we fine-tune the pre-trained model on Ref-Youtube-VOS training set with 1 video sequence per GPU for 6 epochs, where the learning rate is reduced by a factor of 0.1 at the 3th and 5th epoch, respectively. Each video sequence consists of 6 randomly sampled frames from the same video with data augmentations applied, including random horizontal flip, random crop, and photometric distortion. All input frames are resized to have a short side of 360 and the maximum long side of 640. The model is optimized using the AdamW optimizer \cite{loshchilov2018decoupled} with the initial learning rate of $1\times{10}^{-5}$ for the image and text encoders, and $5\times{10}^{-5}$ for the rest parts. It should be noted that the text encoder is optimized during the pre-training phase while its parameters are frozen during the fine-tuning process. Then, the fine-tuned BIFIT model is evaluated on the  Ref-Youtube-VOS and Ref-DAVIS17 val sets. For A2D-Sentences and JHMDB-Sentences datasets, we fine-tune the pre-trained model on the A2D-Sentences training set using the same setting as Ref-Youtube-VOS. Then, we evaluate the performance of the fine-tuned BIFIT model on both the A2D-Sentences test set and JHMDB-Sentences dataset. The loss weights for different losses are set as $\lambda_{cls}=2$, $\lambda_{L1}=5$, $\lambda_{giou}=2$ , $\lambda_{dice}=1$, and $\lambda_{focal}=1$.

\begin{table*}[t]
\caption{Comparison with state-of-the-art methods on the A2D-Sentences test set.}
    \begin{center}
    \tabcolsep 9pt
        \input{table/a2d}
    \end{center}
    \label{A2D}
\end{table*}

\begin{table*}[t]
\caption{Comparison with state-of-the-art methods on the JHMDB-Sentences.}
    \begin{center}
    \tabcolsep 9pt
        \input{table/jhmdb}
    \end{center}
    \label{JHMDB}
\end{table*}

\subsection{Comparison with State-of-the-Art Methods}

\subsubsection{Ref-Youtube-VOS val set} We compare our method with several state-of-the-art approaches on the Ref-Youtube-VOS val set and the results are reported in Table \ref{refytvos}. It can be observed that our BIFIT achieves the overall $J\&F$ accuracy of 59.9\% on the Ref-Youtube-VOS val set, which outperforms the previous methods, such as LBSTI \cite{ding2022language}, MLSA \cite{wu2022multi} and CITD \cite{liang2021rethinking}, with a large marge. Particularly, compared with the ReferFormer with the same ResNet-50 backbone, our BIFIT is \textbf{4.3\%} higher than it at the $J\&F$ accuracy, and even surpasses the ReferFormer with stronger ResNet-101 \cite{he2016deep} backbones by 2.6\%. BIFIT also exceeds the latest work, such as OnlineRefer \cite{OnlineRefer} and R$^2$-VOS \cite{r2vos}. These results validate the superiority of our proposed BIFIT for RVOS task and show that our method achieves state-of-the-art performance.

\subsubsection{ Ref-DAVIS17 val set} We further evaluate the performance of our proposed model on the Ref-DAVIS17 val set. The results are summarized in Table \ref{refdavis}. Following the setting in ReferFormer, we directly report the evaluation results using the model trained on the Ref-Youtube-VOS training set, which means the model is not fine-tuned on the Ref-DAVIS17 dataset. As we can see, our approach realizes the overall $J\&F$ accuracy of 60.5\%, which exceeds all the comparison methods, $e.g.$, BIFIT is 2.0\% and 1.2\% higher than ReferFormer and OnlineRefer, respectively. Moreover, compared with the approaches that are fine-tuned on the Ref-DAVIS17 dataset, such as URVOS \cite{Urvos}, LBSTI \cite{ding2022language} and MLSA \cite{wu2022multi}, our model, without fine-tuning on the target dataset, also outperforms them by a large margin, $e.g.$, 8.9\% higher than URVOS, 6.2\% higher than LBSTI and 2.6\% higher than MLSA, which shows the good generalization of our model. These results further validate the effectiveness of the bidirectional correlation-driven inter-frame interaction Transformer for solving RVOS task.

\subsubsection{A2D-Sentences test set} We further evaluate the performance of the proposed BIFIT on the A2D-Sentences test set and compare it with other cutting-edge methods. The results are presented in Table \ref{A2D}. It can be observed that our approach achieves remarkable 52.4\% mAP, 74.7\% overall IoU and 67.6\% mean IoU, which surpasses all the previous methods. Concretely, compared with the models using the powerful spatial-temporal backbones, our model delivers better results, $e.g.$, a gain of 12.0 mAP over CMPC-V \cite{liu2021cross} with the I3D backbone \cite{i3d} and 7.7 mAP over MTTR \cite{MTTR} using the Video-Swin-T. For the methods with the same ResNet-50 backbone, $e.g.$, LBSTI \cite{ding2022language} and ReferFormer \cite{Referformer}, BIFIT surpasses it in all metrics.

\subsubsection{JHMDB-Sentences} We also assess the performance of our model on JHMDB-Sentences without fine-tuning to further demonstrate the generalization of our approach. As included in Table \ref{JHMDB}, our method beats all the comparison methods on all metrics, except on mAP, which is slightly lower than ClawCraneNet \cite{liang2021clawcranenet} by 0.3\%. Particularly, our method outperforms ReferFormer by 1.2\% and 0.6\% on the mAP and mean IoU, respectively.

\begin{figure*}[ht]
\begin{center}
   \includegraphics[width=0.92\linewidth]{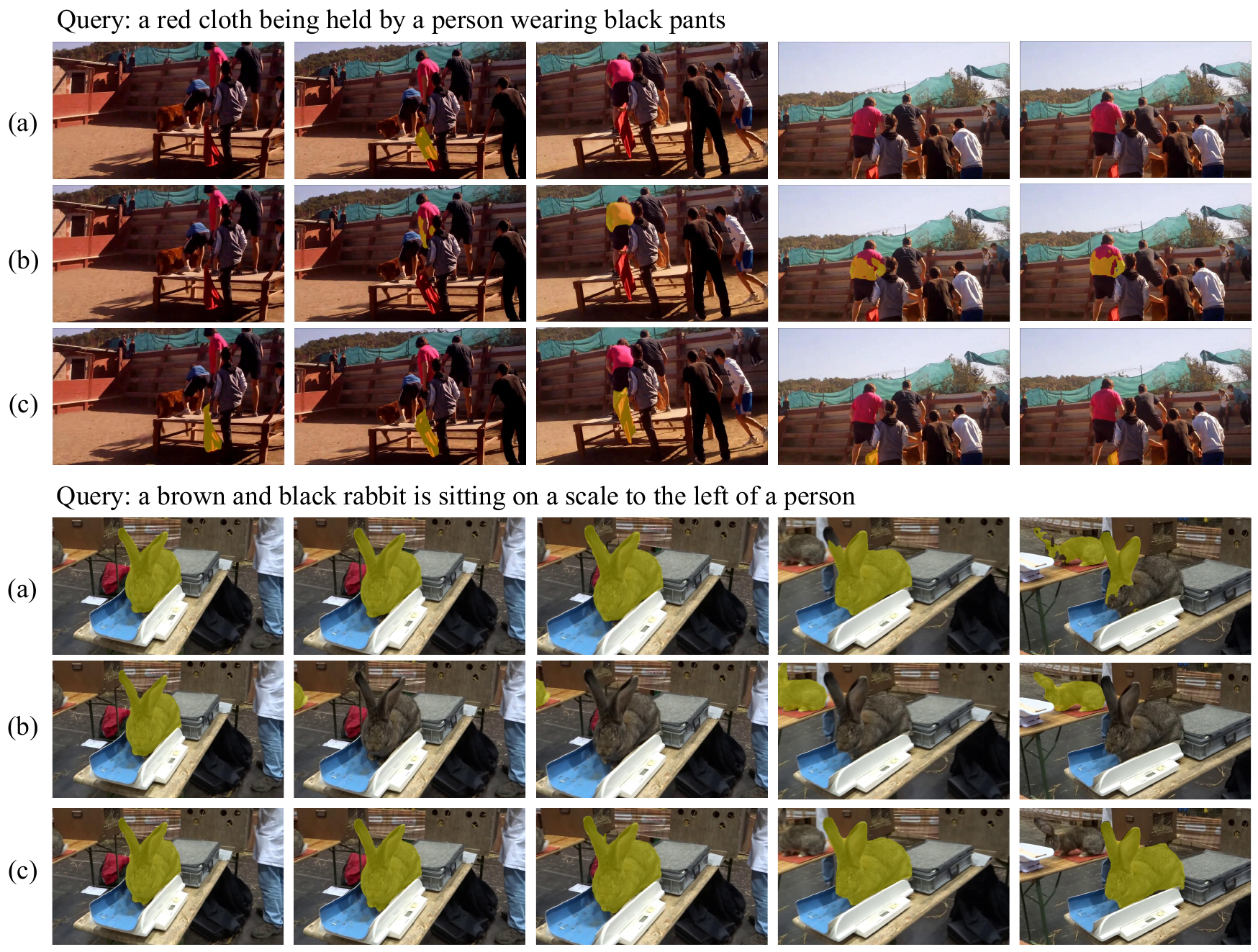}
\end{center}
\caption{Visualization comparison. (a) MTTR \cite{MTTR}. (b) ReferFormer \cite{Referformer}. (c) Our BIFIT.}

\label{vis_com}
\end{figure*}

\begin{figure*}[ht]
\begin{center}
   \includegraphics[width=0.94\linewidth]{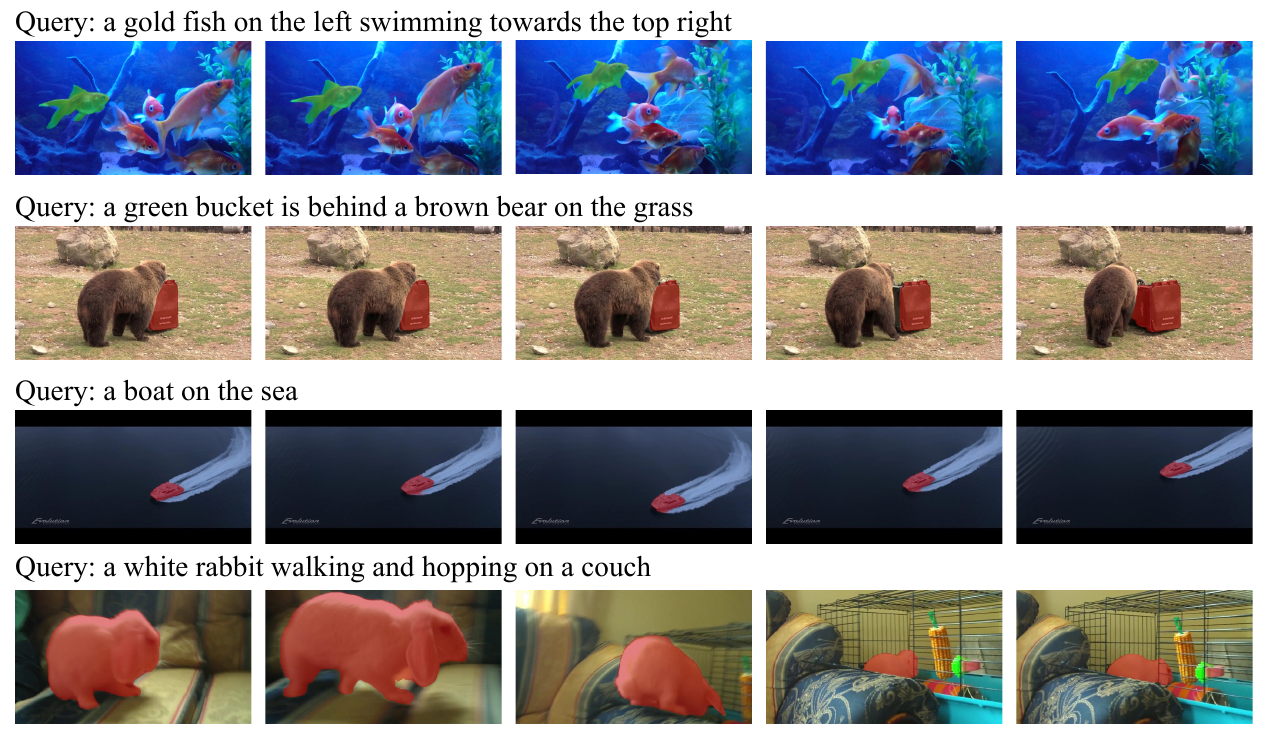}
\end{center}
\caption{Visualization results of our proposed BIFIT on Ref-DAVIS17 and Ref-YouTube-VOS.}
\vspace{-2mm}
\label{vis_res_ours}
\end{figure*}

Fig.\ref{vis_com} presents the visualization comparison between MTTR, ReferFormer and our BIFIT. We can intuitively observe that BIFIT exhibits superior performance compared to MTTR and ReferFormer in terms of the accuracy and consistency of prediction results across frames. Furthermore, we visualize more referring video object segmentation results in Fig.\ref{vis_res_ours}, which demonstrates that BIFIT could cope well with challenging scenarios such as small object, occlusion, and dynamic appearance changes. These results further validate the efficacy of our proposed BIFIT.

\subsection{Model analysis}
In this part, we perform extensive ablation experiments to investigate the influence of the core components of our BIFIT as well as the impacts of different model settings. \textbf{Unless otherwise noted, all of the experiments are conducted on the Ref-Youtube-VOS dataset.}

\begin{table}
\tabcolsep 10pt
\caption{Ablation study of different components of the proposed BIFIT. IFI denotes inter-frame interaction module. BVLIM denotes bidirectional vision-language interaction module.}
    \begin{center}
        \input{table/components}

    \end{center}
    \label{components}
\end{table}

\subsubsection{Components Analysis} To explore the influence of the key components of our model, we first build a baseline model which is the BIFIT without the bidirectional vision-language interaction module and the inter-frame interaction module. As shown in Table \ref{components}, the baseline model obtains an overall $J\&F$ accuracy of 54.0\%. When we add the VEwL and LEwV submodules on the baseline, the $J\&F$ accuracy of baseline model increases to 55.8\% and 55.1\%, respectively. The full bidirectional vision-language interaction module (BVLM) brings a 3.1 $J\&F$ accuracy gains to the baseline. When only inter-frame interaction (IFI) module is integrated to the baseline, $J\&F$ accuracy of the new model increases by 1.3\% to 55.3\%, which is not as significant an improvement as BVLM. The phenomenon can be attributed to the absence of early cross-modal interactions, which results in language queries failing to generate discriminative instance embeddings and consequently impacting the efficacy of the inter-frame interaction module. Therefore, when the VEwL and LEwV submodules are imposed on the baselien with IFI module, the new models improve by 3.5\% and 2.3\% to 58.8\% and 57.6\% $J\&F$ accuracy, respectively. This observation suggests that more discriminative instance embeddings generated in the decoder can enhance the efficacy of the IFI module. Finally, equipped with both proposed modules, our BIFIT realizes the best 59.9\% $J\&F$ accuracy.


\begin{table}
\tabcolsep 9pt
\caption{Model analysis of different settings in the bidirectional vision-language interaction module.}
    \begin{center}
        \input{table/Language_vision_Interaction}
    \end{center}
    \label{VLI}

\end{table}

\subsubsection{Bidirectional Vision-Language Interaction Module} In this study, we explore the impact of different settings in the bidirectional vision-language interaction module. The first is the interaction strategy between visual and text features. We design two types of interaction settings, $i.e.$, 'Attention + Multiply', and 'Attention + FFN'. The former setting is adopted in BIFIT and the later is a common paradigm. The experimental results are reported in Table \ref{VLI}. It can be seen that the BIFIT with the default setting performs better than the one with the later setting, which means that the 'Attention + Multiply' strategy may be more suitable for cross-modal interaction.

The second setting is the interaction procedure of the VEwL submodule. In our model, we use the \textit{fixed} raw text feature to interact with each level of the multi-level visual features. Here we study the another interaction procedure that the next level of the visual feature is interacted with the \textit{dynamically} updated text feature from the last cross-modal interaction. The results are presented in Table \ref{VLI}. It can be observed that BIFIT with the \textit{fixed} text feature setting outperforms the one with dynamical text feature setting by about 2\% $J\&F$ accuracy. The reason may be that the essential information in the raw text feature is not corrupted by the iterative interaction processes, thus making it more suitable for the cross-modal interaction with visual features.

\begin{figure}[t]
    \centering
     \includegraphics[width=0.92\linewidth]{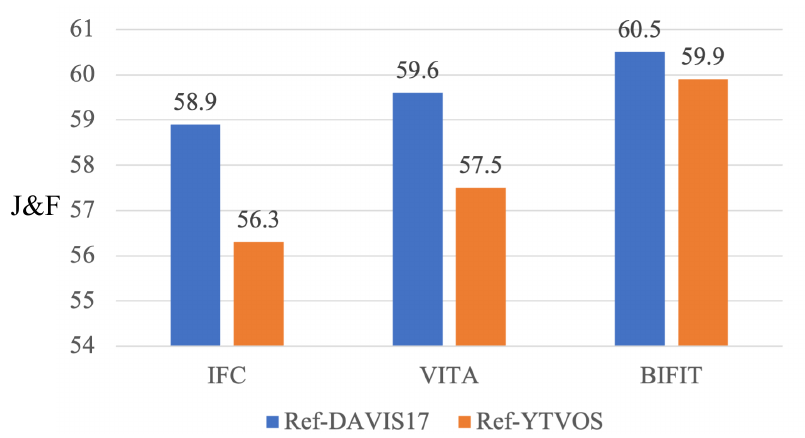}
     \caption{Ablation study of the superiority of IFI module on Ref-DAVIS17 and Ref-YoutubeVOS.}
     \vspace{-2mm}
     \label{ifi_ex}
\end{figure}

\begin{table}
\caption{Ablation study of the generalization of IFI module.}
    \begin{center}
    \tabcolsep 8pt
        \begin{tabular}{l | c | c c c }
        \toprule
        {Method} & Backbone & $\mathcal{J}\&\mathcal{F}$ & $\mathcal{J}$ & $\mathcal{F}$ \\
        \midrule
        MTTR ($w$=8) & Video-Swin-T &53.0 & 51.4 & 54.6 \\
        MTTR ($w$=8) + \textbf{IFI} & Video-Swin-T & \textbf{54.0} & \textbf{52.5} & \textbf{55.5} \\
        \midrule
        ReferFormer & ResNet-50 & 55.6 & 54.8 & 56.5 \\
        ReferFormer + \textbf{IFI} & ResNet-50 & \textbf{58.4} & \textbf{57.0} & \textbf{59.8} \\
        \bottomrule
        \end{tabular}
\vspace{-3mm}
    \end{center}
    \label{ifi_ge}
\end{table}

\subsubsection{Inter-frame Interaction Module} Here, we investigate the superiority and generalization of the inter-frame interaction module. First, we examine the superiority of our IFI module on Ref-DAVIS17 and  Ref-YoutubeVOS val sets by comparing it with the strategies in IFC \cite{ifc} and VITA \cite{vita}. For IFC, we replace the inter-frame interaction Transformer of our BIFIT with the inter-frame communication Transformers in IFC to achieve temporal correlation. For VITA, we remove the IFI module and attach the Transformer in VITA behind the multimodal Transformer of BIFIT to associate frame-level object tokens. The results are presented in Fig.\ref{ifi_ex}. It can be observed that the BIFIT with our IFI moudle achieves the best performance on both datasets, which verifies the superiority of the IFI module. Second, to evaluate the generalization of the IFI module, we apply it to the decoders of other multimodal Transformer based ROVS methods, $i.e.$, MTTR \cite{MTTR} and ReferFormer \cite{Referformer}, and keep all the other settings unchanged. The results are reported in Table \ref{ifi_ge}. As we can see that IFI module brings 1.0\% $J\&F$ accuracy for MTTR and 2.8\% $J\&F$ accuracy for ReferFormer, thereby substantiating the generalization of the IFI module.

\begin{table}[ht]
\caption{Ablation study of the number of inter-frame interaction layers. Ratio indicates the ratio of the numbers of the decoder layers and the subsequent interaction layers.}
\vspace{-2mm}
    \begin{center}
    \tabcolsep 12pt
        \begin{tabular}{l | c | c c c }
        \toprule
        {Method} & {Ratio} & $\mathcal{J}\&\mathcal{F}$ & $\mathcal{J}$ & $\mathcal{F}$ \\
        \midrule
        BIFIT & 1:2 & 58.2 & 56.8 & 59.6 \\
        BIFIT & 2:1 & 57.1 & 55.7 & 58.4 \\
        BIFIT & 1:1 & \textbf{59.9} & \textbf{58.4} & \textbf{61.4} \\
        \bottomrule
        \end{tabular}
\vspace{-2mm}        
    \end{center}
    \label{num_inter-frame-layer}
\end{table}

\subsubsection{Number of Inter-frame Interaction Layers} In this part, we examine the effects of varying numbers of inter-frame interaction layers that follow each decoder layer. Specifically, we assess the performance of BIFIT with varying ratios, which means the ratio of the number of decoder layers and subsequent interaction layers. We perform the experiments with ratios of 1:2, 2:1 and 1:1, and the results are reported in Table \ref{num_inter-frame-layer}. As we can observe that when the ratio is 1:1, the BIFIT model could achieve the best performance and significantly exceeds the models with other ratio settings. Thus we use the ratio of 1:1 as the default setting of BIFIT.

\begin{figure}[t]
    \centering
     \includegraphics[width=0.96\linewidth]{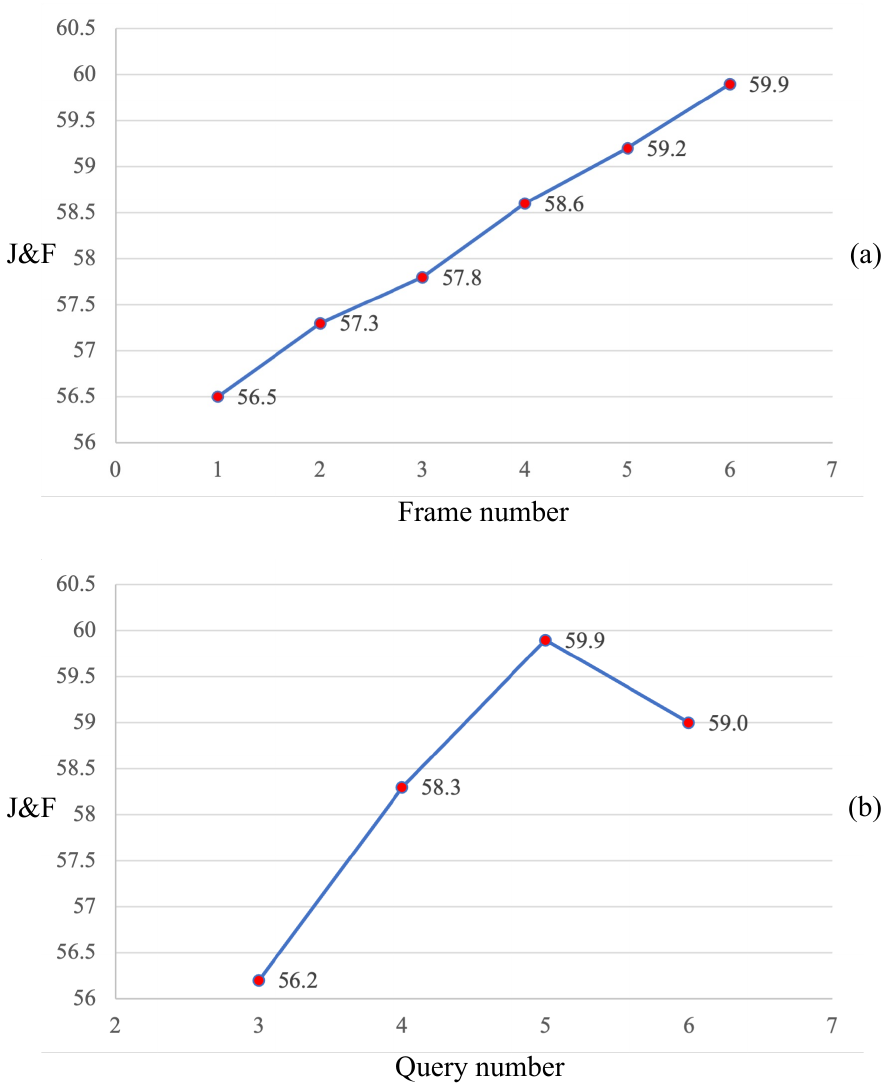}
     \caption{Ablation studies of the number of training frames and the number of the query embeddings.}
     \vspace{-2mm}
     \label{num_frame_query}
\end{figure}

\subsubsection{Number of Training Frames} Here, we explore the influence of the number of the training frames during the fine-tuning stage on the model performance. The experiments were conducted using 4, 5, and 6 training frames, respectively. The results are presented in Fig. \ref{num_frame_query} (a). It can be seen that as the number of training frames increases, the performance of the model improves gradually, and we get the best performance of 59.9\% $\mathcal{J}\&\mathcal{F}$ accuracy using 6 training frames during training process. It is worth noting that the maximum number of training frames we can set is limited to 6 due to constraints on our GPU memory. Therefore, increasing the number of frames may lead to further improvement in model accuracy.


\subsubsection{Number of language queries} In this study, we investigate the impact of varying numbers of language queries on model performance. We conduct experiments with language queries number of 4, 5 and 6, respectively. The experimental results are depicted in Fig. \ref{num_frame_query} (b). We can observe that the BIFIT model attains the best accuracy when using 5 language queries, and more language queries, $e.g.$, 6 language queries, can even degrade the performance. Therefore, we choose $N=5$ as the number of language queries for BIFIT.


\section{Conclusion}

In this paper, we propose a bidirectional correlation-driven inter-frame interaction Transformer, termed BIFIT, to solve the issues of inter-frame interaction and cross-modal correlation in the multimodal Transformer based RVOS methods. We design the lightweight and effective inter-frame interaction module and insert it into the multimodal Transformer decoder to efficiently model the temporal coherence and learn the spatio-temporal representation of the referred object, so as to decode more consistent instance embeddings for predicting high-quality segmentation results. Moreover, we devise the bidirectional vision-language interaction module and place it before the inter-frame interaction Transformer to enhance the correlation between the cross-modal features, thus facilitating the language queries to decode more precise object information from visual features and further improving the model performance. Experimental results on four benchmarks validate the superiority of our BIFIT over state-of-the-art methods and the effectiveness of our proposed modules.

\bibliography{main}

\end{document}

%% file: table/refytvos.tex
\begin{tabular}{l | c | c c c }
\toprule
Method & Backbone & $\mathcal{J}\&\mathcal{F}$ & $\mathcal{J}$ & $\mathcal{F}$ \\
\midrule
URVOS ~\cite{Urvos} & ResNet-50 & 47.2 & 45.3 & 49.2 \\
YOFO ~\cite{yofo} & ResNet-50 & 48.6 & 47.5 & 49.7 \\
LBSTI ~\cite{ding2022language} & ResNet-50 & 49.4 & 48.2 & 50.6 \\
MLSA ~\cite{wu2022multi} & ResNet-50 & 49.7 & 48.4 & 51.0 \\
MTTR \cite{MTTR} & Video-Swin-T & 55.3 & 54.0 & 56.6 \\
ReferFormer ~\cite{Referformer} & ResNet-50 & 55.6 & 54.8 & 56.5 \\
CITD \cite{liang2021rethinking} & ResNet-101 & 56.4 & 54.8 & 58.1 \\
OnlineRefer \cite{OnlineRefer} & ResNet-50 & 57.3 & 55.6 & 58.9 \\
R$^2$-VOS \cite{r2vos} & ResNet-50 & 57.3 & 56.1 & 58.4 \\
ReferFormer ~\cite{Referformer} & ResNet-101 & 57.3 & 56.1 & 58.4 \\
BIFIT & ResNet-50 & \textbf{59.9} & \textbf{58.4} & \textbf{61.4} \\ 

\bottomrule

\end{tabular}

%% file: table/davis.tex
\begin{tabular}{l | c | c c c }

\toprule

\multirow{1}{*}{Method} & \multirow{1}{*}{Backbone} 
& $\mathcal{J}\&\mathcal{F}$ & $\mathcal{J}$ & $\mathcal{F}$ \\

\midrule
CMSA+RNN ~\cite{CMSA} & ResNet-50 & 40.2 & 36.9 & 43.5 \\
URVOS ~\cite{Urvos} & ResNet-50 & 51.6 & 47.3 & 56.0 \\
MLSA$^{*}$ ~\cite{wu2022multi} & ResNet-50 & 52.7 & 50.0 & 55.4 \\
LBSTI ~\cite{ding2022language} & ResNet-50 & 54.3 & - & - \\
MLSA ~\cite{wu2022multi} & ResNet-50 & 57.9 & 53.9 & 62.0 \\
ReferFormer$^{*}$ ~\cite{Referformer} & ResNet-50 & 58.5 & 55.8 & 61.3 \\
OnlineRefer$^{*}$ \cite{OnlineRefer} & ResNet-50 & 59.3 & 55.7 & 62.9 \\
R$^2$-VOS$^{*}$ \cite{r2vos} & ResNet-50 & 59.7 & 57.2 & 62.4 \\
BIFIT$^{*}$ & ResNet-50 & \textbf{60.5} & \textbf{56.9} & \textbf{64.1} \\

\bottomrule

\end{tabular}

%% file: table/a2d.tex
\begin{tabular}{l | c | c c c c c | c c | c }
\toprule

\multirow{2}{*}{Method} & \multirow{2}{*}{Backbone} & \multicolumn{5}{c |}{Precision} & \multicolumn{2}{c |}{IoU} & \multirow{2}{*}{mAP} \\

 & & P@0.5 & P@0.6 & P@0.7 & P@0.8 & P@0.9 & Overall & Mean &  \\
\midrule
Hu $et al.$ \cite{hu2016segmentation} & VGG-16 & 34.8 & 23.6 & 13.3 & 3.3 & 0.1 & 47.4 & 35.0 & 13.2 \\
ACAN ~\cite{wang2019asymmetric} & I3D & 55.7 & 45.9 & 31.9 & 16.0 & 2.0 & 60.1 & 49.0 & 27.4 \\
CSTM ~\cite{hui2021collaborative} & I3D & 65.4 & 58.9 & 49.7 & 33.3 & 9.1 & 66.2 & 56.1 & 39.9 \\
CMPC-V ~\cite{liu2021cross} & I3D & 65.5 & 59.2 & 50.6 & 34.2 & 9.8 & 65.3 & 57.3 & 40.4 \\
MTTR ~\cite{MTTR} & Video-Swin-T & 72.1 & 68.4 & 60.7 & 45.6 & 16.4 & 70.2 & 61.8 & 44.7 \\
LBSTI ~\cite{ding2022language} & ResNet-50 & 73.0 & 67.4 & 59.0 & 42.1 & 13.2 & 70.4 & 62.1 & 47.2 \\
ClawCraneNet \cite{liang2021clawcranenet} & ResNet-50/101 & 70.4 & 67.7 & 61.7 & 48.9 & 17.1 & 63.1 & 59.9 & 49.4 \\
ReferFormer ~\cite{Referformer} & ResNet-50 & 78.9 & 75.7 & 68.7 & 51.5 & 17.6 & 74.5 & 66.5 & 51.1 \\
BIFIT & ResNet-50 & \textbf{80.0} & \textbf{77.2} & \textbf{70.2} & \textbf{53.6} & \textbf{19.8} & \textbf{74.7} & \textbf{67.6} & \textbf{52.4} \\

\bottomrule
\end{tabular}
\vspace{-3mm}

%% file: table/jhmdb.tex
\begin{tabular}{l | c | c c c c c | c c | c }

\toprule

\multirow{2}{*}{Method} & \multirow{2}{*}{Backbone} & \multicolumn{5}{c |}{Precision} & \multicolumn{2}{c |}{IoU} & \multirow{2}{*}{mAP} \\

 & & P@0.5 & P@0.6 & P@0.7 & P@0.8 & P@0.9 & Overall & Mean &  \\
\midrule
Hu $et al.$ \cite{hu2016segmentation} & VGG-16 & 63.3 & 35.0 & 8.5 & 0.2 & 0.0 & 54.6 & 52.8 & 17.8 \\
ACAN ~\cite{wang2019asymmetric} & I3D & 75.6 & 56.4 & 28.7 & 3.4 & 0.0 & 57.6 & 58.4 & 28.9 \\
CSTM ~\cite{hui2021collaborative} & I3D & 78.3 & 63.9 & 37.8 & 7.6 & 0.0 & 59.8 & 60.4 & 33.5 \\
CMPC-V ~\cite{liu2021cross} & I3D & 81.3 & 65.7 & 37.1 & 7.0 & 0.0 & 61.6 & 61.7 & 34.2 \\
MTTR ~\cite{MTTR} & Video-Swin-T & 91.0 & 81.5 & 57.0 & 14.4 & 0.1 & 67.4 & 67.9 & 36.6 \\
LBSTI ~\cite{ding2022language} & ResNet-50 & 86.4 & 74.4 & 53.3 & 13.2 & 0.0 & 64.5 & 65.8 & 41.1 \\
ClawCraneNet \cite{liang2021clawcranenet} & ResNet-50/101 & 88.0 & 79.6 & 56.6 & 14.7 & 0.2 & 64.4 & 65.6 & \textbf{43.3} \\
ReferFormer ~\cite{Referformer} & ResNet-50 & 95.6 & 87.8 & 65.4 & 19.0 & 0.2 & 72.0 & 70.9 & 41.8 \\
BIFIT & ResNet-50 & \textbf{95.8} & \textbf{88.2} & \textbf{66.5} & \textbf{20.2} & \textbf{0.3} & \textbf{72.1} & \textbf{71.5} & 43.0 \\

\bottomrule
\end{tabular}
\vspace{-2mm}

%% file: table/components.tex





\tabcolsep 8pt 
\begin{tabular}{c|ll|l|lll}
\toprule
\multicolumn{1}{c|}{\multirow{2}{*}{Method}} & \multicolumn{2}{c|}{BVLIM} & \multirow{2}{*}{IFI} & \multicolumn{1}{c}{\multirow{2}{*}{$\mathcal{J}\&\mathcal{F}$}} & \multicolumn{1}{c}{\multirow{2}{*}{$\mathcal{J}$}} & \multicolumn{1}{c}{\multirow{2}{*}{$\mathcal{F}$}} \\
\multicolumn{1}{c|}{} & VEwL & LEwV &  & \multicolumn{1}{c}{} & \multicolumn{1}{c}{} & \multicolumn{1}{c}{} \\ 
\midrule
Baseline &   &   &    &   54.0   &  52.9  &  55.1 \\
BIFIT & \checkmark  &   &    & 55.8  &   55.0 & 56.6 \\
BIFIT &    & \checkmark &    & 55.1 & 53.7 & 56.5 \\
BIFIT &  \checkmark  & \checkmark &    &  57.1 & 55.9 & 58.3 \\
\midrule
BIFIT  &   &      &  \checkmark   &  55.3   &  54.4  &  56.3   \\ 
BIFIT & \checkmark  &   &  \checkmark  & 58.8 & 57.4 & 60.2  \\
BIFIT &    & \checkmark &  \checkmark  & 57.6 & 56.4 & 58.7 \\
\midrule
BIFIT  &  \checkmark &   \checkmark   &  \checkmark   &\textbf{59.9} & \textbf{58.4} & \textbf{61.4} \\
\bottomrule

\end{tabular}
\vspace{-4mm}

%% file: table/Language_vision_Interaction.tex
\begin{tabular}{l | c | c c c }

\toprule

{Method} & {Settings} & $\mathcal{J}\&\mathcal{F}$ & $\mathcal{J}$ & $\mathcal{F}$ \\

\midrule
\multicolumn{5}{l}{\textbf{Interaction Strategies}} \\
\midrule

BIFIT & Attention+Multiply & \textbf{59.9} & \textbf{58.4} & \textbf{61.4} \\
BIFIT & Attention+FFN & 58.7 & 57.4 & 60.1 \\
\midrule
\multicolumn{5}{l}{\textbf{Interaction Procedure of VEwL submodule}} \\
\midrule
BIFIT & Fixed & \textbf{59.9} & \textbf{58.4} & \textbf{61.4} \\
BIFIT & Dynamic & 58.0 & 56.7 & 59.3 \\

\bottomrule
\end{tabular}
\vspace{-2mm}